\documentclass[sigconf]{acmart}

\usepackage{amsmath}
\usepackage{algpseudocode}
\usepackage[ruled,linesnumbered]{algorithm2e}
\usepackage{amsfonts}
\usepackage{verbatimbox}
\usepackage{hyperref}

\AtBeginDocument{%
  }

\begin{document}

\title{Enhancing the Resilience of Graph Neural Networks to Topological Perturbations in Sparse Graphs}

\author{Shuqi He\(^1\), Jun Zhuang\(^2\), Ding Wang\(^1\), Luyao Peng\(^1\), Jun Song\(^1\)}
\affiliation{
  \(^1\) China University of Geosciences (Wuhan), Wuhan, China \\
  \(^2\) Indiana University-Purdue University Indianapolis, Indianapolis, USA \\
  \country{}
  {\small \{1202211192, wangding, pengluyao, songjun\}@cug.edu.cn, junz@iu.edu}
}

\begin{abstract}
Graph neural networks (GNNs) have been extensively employed in node classification.   Nevertheless, recent studies indicate that GNNs are vulnerable to topological perturbations, such as adversarial attacks and edge disruptions. Considerable efforts have been devoted to mitigating these challenges. For example, pioneering Bayesian methodologies, including GraphSS and LlnDT, incorporate Bayesian label transitions and topology-based label sampling to strengthen the robustness of GNNs. However, GraphSS is hindered by slow convergence, while LlnDT faces challenges in sparse graphs. To overcome these limitations, we propose a novel label inference framework, TraTopo, which combines topology-driven label propagation, Bayesian label transitions, and link analysis via random walks. TraTopo significantly surpasses its predecessors on sparse graphs by utilizing random walk sampling, specifically targeting isolated nodes for link prediction, thus enhancing its effectiveness in topological sampling contexts. Additionally, TraTopo employs a shortest-path strategy to refine link prediction, thereby reducing predictive overhead and improving label inference accuracy. Empirical evaluations highlight TraTopo's superiority in node classification, significantly exceeding contemporary GCN models in accuracy.
\end{abstract}


\maketitle

\section{Introduction}
Graph structures, such as attributed graphs~\cite{tian2023heterogeneous, zhuang2024robust}, knowledge graphs~\cite{liu2022joint, liu2021neural}, and factor graphs~\cite{chen2022multi, chen2022multi2, chen2023learning}, play a crucial role across various domains, representing the topological relationships and attribute information between nodes. Node classification is a fundamental task in graph structure learning. In this task, we aim to assign the nodes to the corresponding class.

In recent years, Graph Neural Networks (GNNs) have been widely applied in node classification due to their superior performance on graph representation~\cite{tian2022learning, tian2024graph, chen2022scalable, wu2022graph, tian2022reciperec, hamilton2017inductive, velivckovic2017graph, zhuang2024understanding}. However, recent studies reveal that GNNs may be vulnerable to topological perturbations, which can severely compromise the effectiveness of GNN-based node classification~\cite{liu2021topological, ding2024toward, zhang2020topological}. Thus, it is crucial to improve the robustness of GNNs against topological perturbations, such as random perturbations~\cite{zhuang2022deperturbation, hahn2020random} and graph sparsification~\cite{fan2021sparse, zhuang2023robust}.

Numerous studies, such as Bayesian Label Transition~\cite{yang2022estimating} and label propagation~\cite{cordasco2012label}, have been explored to improve the robustness of GNNs. These approaches adeptly utilize supervised data to enhance robustness, yet the effectiveness is circumscribed by the inherent characteristics of local graph structures, which may inhibit the propagation process for unlabeled nodes. GraphSS~\cite{zhuang2022defending} endeavors to counteract suboptimal classification outcomes stemming from topological perturbations by refining Graph Neural Network (GNN) predictions through post-processing. This strategy incorporates a Bayesian inference framework to devise a label transition matrix, thereby substituting misjudged labels with more accurate alternatives to ameliorate classification discrepancies. Nonetheless, the adaptation of this technique is hampered by its protracted convergence rate. A novel initiative, LInDT~\cite{zhuang2022robust}, addresses the challenge of delayed convergence by introducing an innovative label sampling technique, thereby enhancing the method's scalability across expansive graph structures. Despite these advancements, LInDT's dependence on the underlying graph topology renders it less effective on sparsely connected graphs, where limited connectivity can severely diminish the success of label propagation.

To address the aforementioned challenges, we introduce a novel mechanism, namely TraTopo, which integrates Random Walk with Restart and PageRank algorithms to augment the robustness of topology-based propagation methodologies. This model is seamlessly integrated within a Bayesian label transition framework, thus strengthening the resilience of GNNs in node classification tasks.
More precisely, TraTopo outperforms its predecessor, LlnDT, by employing label propagation to achieve enhanced convergence in scenarios of uncertain Bayesian label sampling. It leverages random walk-based algorithms to adeptly navigate the constraints presented by nodes of lower degrees, while concurrently diminishing computational burdens. The mechanism we propose not only enriches node information but also refines label inference capabilities, thereby manifesting exemplary performance across graph datasets under conditions of perturbation.
In the experiments, we evaluate the performance of TraTopo and comparative models in terms of accuracy and entropy under a range of topological perturbations across three public datasets. Besides, we analyze the sensitivity of various hyper-parameters in TraTopo. Our systematic validation seeks to enhance the robustness and the predictive capabilities of TraTopo in dynamic and diverse structural graph data.
Overall, our main contributions are summarized as follows:
\begin{itemize}
    \item We propose a new mechanism for node label inference by integrating Bayesian methods with topology-based enhancements, incorporating Random Walk with Restart and PageRank to boost link prediction accuracy.
    \item We employ shortest-path-based strategies to streamline random walks, reducing computational overhead and enhancing predictive performance with minimal resource consumption.
    \item Extensive experiments demonstrate that our method can outperform leading competing models across benchmark graph datasets, validating the effectiveness in dynamic network environments.
\end{itemize}

\section{RELATED WORK}
\par Node classification is crucial in analyzing graph-structured data for social networks, bioinformatics, and recommendation systems. Advances in this field include Graph Neural Networks (GNNs), adversarial robustness, noisy label management, and algorithms like random walk and PageRank.
\subsection{Graph Neural Networks}
\par Graph Neural Networks (GNNs) are essential for analyzing graph-structured data, aiding in areas such as social network analysis, bioinformatics, and recommendation systems. A major challenge is maintaining GNN robustness against accidental or adversarial topology perturbations.
\par Recent studies have explored black-box adversarial attacks on Graph Neural Networks (GNNs), employing a node voting strategy to identify vulnerable nodes~\cite{wen2024black}. Fiorellino et al.~\cite{fiorellino2024topological} have developed an advanced GNN variant designed to enhance resilience against channel perturbations. Furthermore, Khalid et al.~\cite{khalid2024sleepnet} introduced SleepNet, an innovative sleep prediction model that incorporates attention mechanisms and utilizes dynamic social networks.

\subsection{Adversarial Robustness}
\par With the rise of Graph Neural Networks (GNNs), their susceptibility to adversarial tactics has captured academic focus~\cite{zhang2023assessing, xu2024enhancing}. Research prioritizes bolstering network security through tailored attacks and enhanced defenses. Notably, even minimal, strategic perturbations substantially reduce the efficacy of GNNs, challenging their precision and interpretability.
\par Zhao et al.~\cite{zhao2024adversarial} employed a Hamiltonian method to enhance GNN resilience against topological disturbances, elevating stability across GNN architectures. Wu et al.~\cite{wu2023adversarial} improved GCN robustness and generalization via weight perturbations, noting that optimizing robust loss directly enhances defenses. Liu et al.~\cite{liu2024towards} introduced wave-induced resonance to boost GNN robustness. Testa et al.~\cite{testa2024stability} analyzed GNN stability via slight perturbations. Liu et al.~\cite{liu2024revisiting} examined the impact of edge perturbations on GNN robustness and vulnerability.

\subsection{Noisy Labels}
\par Learning with noisy labels substantially alters training dynamics, potentially reducing model performance~\cite{zhang2021noilin, chen2023uncertainty, tian2022nosmog}. In node classification, structural dependencies in graphs exacerbate inaccuracies, facilitating the spread of incorrect labels through connecting edges.

\par Zhang et al.~\cite{zhang2024badlabel} devised an advanced LNL algorithm to effectively address noisy labels. Xia et al.~\cite{xia2023gnn} introduced a GNN-based Cleaner, enhancing robustness against noisy labels in attributed graphs. Self-supervised methods have become pivotal in graph representation learning~\cite{zhai2023understanding}. Zhuang et al.~\cite{zhuang2022does} pioneered the concept of treating noisy labels as intrinsic data properties. Yuan et al.~\cite{yuan2023self} developed a self-supervised framework designed to mitigate the impact of noisy graphs and labels. 

\subsection{Random Walk and PageRank}
\par Graph Convolutional Networks (GCNs) tackle structural disruptions using advanced random walk and PageRank, enhancing resilience and efficiency across various graph-learning contexts.
\par Utilizing APPNP's~\cite{gasteiger2018predict} Personalized PageRank and N-GCN's~\cite{abu2020n} stochastic walks bolsters GCN resilience, streamlining topological coherence and nodal comprehension. Wang et al.~\cite{wang2023scapin} advocate robustness assessments through graph perturbations, underscoring diffusion and influence maximization's defensive prowess. Hou et al.~\cite{hou2023can} probe directed graph resilience via BBRW, spotlighting the fortifying influence of targeted pathways, and advancing graph topology understanding.

\section{PRELIMINARIES}
In this section, we introduce the preliminary background about GNNs and random walks.

\subsection{GNN-based Node Classification}
\par In this investigation, we employ Graph Convolutional Networks (GCNs)~\cite{yao2019graph} as the foundational \textbf{node classifier} \(f_\theta\), constructing an undirected, attributed \textbf{graph} \(G = (V, E)\) composed of \(N\) vertices and corresponding edges. The structure is defined by a \textbf{symmetric adjacency matrix} \(A\) and a \textbf{feature matrix} \(X\), formally expressed as \(A \in \mathbb{R}^{N \times N}\) and \(X \in \mathbb{R}^{N \times d}\), respectively.

\par Graph Convolutional Networks (GCNs) have gained prominence for their capability to perform convolution operations on graph-structured data. The fundamental operation of a GCN can be described by the layer-wise propagation rule:
\begin{equation}
    H^{(l+1)} = \sigma\left( \tilde{D}^{-1/2} \tilde{A} \tilde{D}^{-1/2} H^{(l)} W^{(l)} \right)
\end{equation}
where \(\tilde{A} = A + I\) is the adjacency matrix \(A\) of the graph with added self-loops, \(\tilde{D}\) is the corresponding degree matrix, \(H^{(l)}\) denotes the matrix of activations in the \(l\)-th layer (with \(H^{(0)} = X\)), \(W^{(l)}\) is the matrix of trainable weights in the \(l\)-th layer, and \(\sigma\) is a nonlinear activation function. This formula captures the essence of GCNs in aggregating features from a node’s local neighborhood, thereby enabling the model to learn powerful representations from graph-structured inputs.

\par Utilizing \(A\) and \(X\) for the task of node classification, we integrate noisy labels as a sophisticated regularization mechanism aimed at bolstering the model's resilience against data imbued with noise. Specifically, a subset of nodes, denoted as \(\mathcal{U}\) and comprising 10\% of the graph's total, is assigned \textbf{noisy labels} \(\mathbf{Y}\). These labels are an amalgamation of \textbf{manually-annotated labels} \(\mathbf{Y}_m\) and \textbf{auto-generated labels} \(\mathbf{Y}_a\). This methodology corroborates that an elevation in noise levels can substantially augment the efficacy of the regularization process. Furthermore, our scholarly objective is to meticulously align the \textbf{inferred labels} \(\hat{\mathbf{Z}}\) as closely as possible with the \textbf{latent labels} \(\mathbf{Z}\), thus ensuring robust node classification within noisy environments. This approach not only demonstrates the feasibility of effectively leveraging graph-structured data in complex labeling landscapes but also delves into how advanced regularization techniques can significantly enhance the model’s ability to adapt to noise and improve its overall performance.

\subsection{Random walk algorithm}
\par 
The Random Walk algorithm is a stochastic graph traversal method that simulates the process of moving randomly within a graph. This algorithm finds extensive applications in graph data, including network analysis, link analysis, and ranking of graph nodes.
\subsubsection{Random walk with restart}
\par The Random Walk with Restart (RWR) The algorithm~\cite{xia2019random,li2015random,tong2006fast} refines personalized exploration within network analysis, optimizing the evaluation of node importance and the disclosure of subgraphs.
The algorithm operates on a probabilistic mechanism: returning to the starting node with probability $\alpha$ or advancing to a neighbor with $1-\alpha$. The matrix $P$ underlies the RWR update: \begin{equation} RWR(v, t+1) = \alpha \cdot RWR(v, t) + (1 - \alpha) \sum_{u \in \mathcal{N}(v)} \frac{RWR(u, t)}{\text{deg}(u)} \end{equation} Guided by $\alpha$, the RWR algorithm performs a stochastic traversal of the graph.
\subsubsection{PageRank}
PageRank~\cite{xing2004weighted}, used by Google, ranks web pages based on link importance. It calculates the rank \( PR(A) \) using:
\begin{equation}
    PR(A) = (1-d) + d \left( \sum_{i=1}^n \frac{PR(T_i)}{C(T_i)} \right)
\end{equation}
where \( d \) (typically 0.85) is the damping factor, \( T_i \) are linking pages, \( PR(T_i) \) is their PageRank, and \( C(T_i) \) is their outbound links. This iterative method ranks pages by link structure.
\section{METHOD}
\par This section presents TraTopo, combining Bayesian label propagation with ensemble learning to improve link prediction and reduce errors. It employs a shortest-path algorithm to identify new nodes, update candidates, and lower computational demands.

\subsection{Bayesian Label Transition with Asymmetric Dirichlet Distributions}
Using Bayesian theory, the Bayesian Label Propagation algorithm estimates nodes' label probability distributions ~\cite{yang2022estimating, liu2023identifiability, xie2013labelrank, zhuang2022robust}. It calculates likelihoods from neighboring labels, represents initial distributions with prior probabilities, and iteratively refines these distributions to enhance label propagation.

\par The algorithm initializes by establishing an initial label probability distribution per node, subsequently refined through iterative updates informed by adjacent nodes and propagation protocols. Bayesian adjustments recalibrate the probabilities of nodes with known labels. This iterative refinement proceeds until stabilization or a designated iteration threshold is met. The final label distribution for a node $v$ is represented by $L_{v}$.

\begin{equation}
P(L_v = l \mid \text{Neib}(v), Y) \propto \sum_{u \in \text{Neib}(v)} P(L_u = l \mid Y)
\end{equation}
where \textbf{Neighbors} $Neib(v)$ represents the adjacent nodes of $v$, and \textbf{observed labels} $Y$ denote known label information. The node's label probability distribution is updated using Bayesian inference, where 
\(P(L_u=l \mid Y)\) indicates the probability that node $u$ has label $l$ based on observed label information. The label propagation process iterates these updates until convergence.

\par The Bayesian label transition utilized in this study is illustrated in Figure 1.

\begin{figure}[h]
  \centering
  \includegraphics[width=0.8\linewidth]{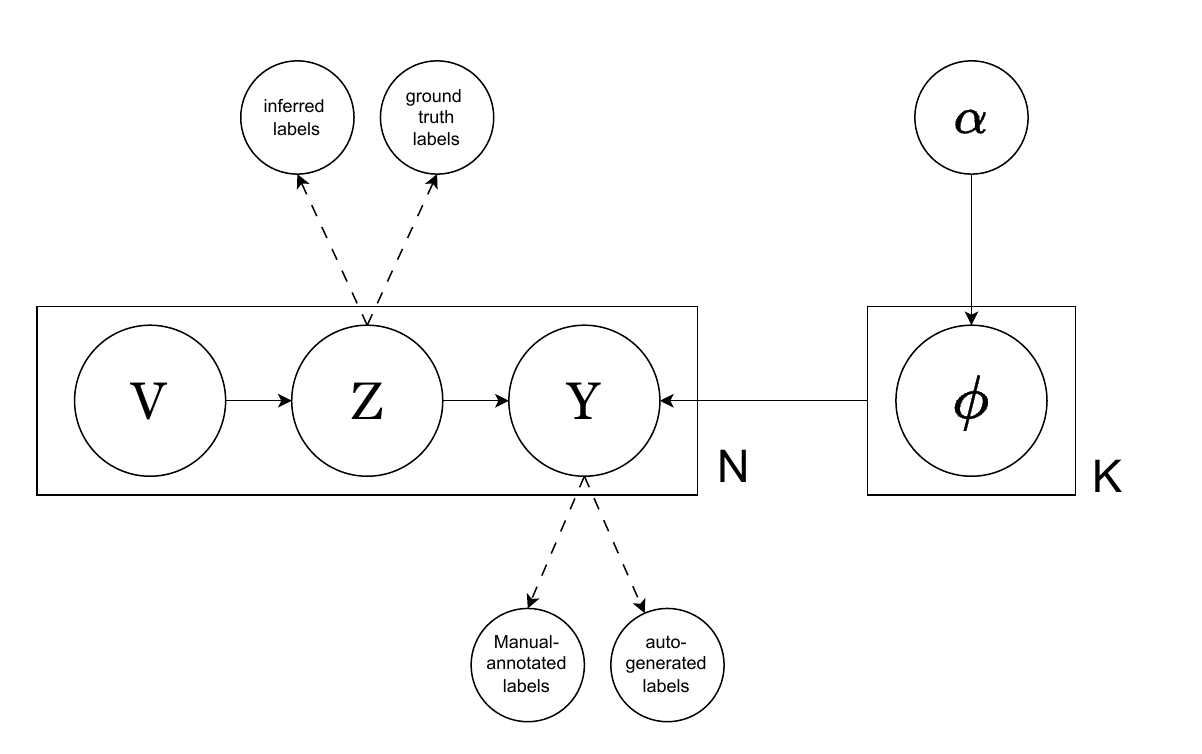}
  \caption{The diagram of Bayesian label transition, $V$ signifies nodes and $N$ indicates the number of nodes. $Z$ includes inferred $\bar{\mathcal{Z}}$ and true labels $Z$, and $Y$ encompasses both manually-annotated labels $Y_m$ and automatically-generated labels $Y_a$ labels. The $K$ class label transition, controlled by matrix $\phi$ and parameter $\alpha$, Black arrows depict variable dependencies, while dotted arrows indicate this symbol can be subdivided into two different meanings.}
  \label{graph:bayesian}
\end{figure}
In the diagram depicted in~~\ref{graph:bayesian}, foundational elements—vertices ($V$), latent labels ($Z$), and noisy labels ($Y$)—are crucial for deciphering the model’s architecture and function. Vertices ($V$) indicate the nodes, latent labels ($Z$) are characterized both as transitionally inferred and true labels, while noisy labels ($Y$) are differentiated into manually annotated and automatically-generated labels. The principal goal is ensuring that the inferred labels ($\bar{Z}$) are in precise concordance with the true labels. Solid arrows signify dependencies, and dashed arrows indicate that there are two definitions for this element.This matrix, parameterized by $\alpha$, governs label transitions, represented as $\phi = [\phi_1, \phi_2, ..., \phi_K]^T \in \mathbb{R}^{K \times K}$, containing $K$ vectors. Each vector $\phi_k$ originates from an Asymmetric Dirichlet Distribution $\phi(\alpha_k)$. The model dynamically revises $\alpha$. For example, $\alpha_k^t$ during the $t$th transition is expressed as
\begin{equation}
\alpha_k^t = \alpha_k^{t-1} \frac{\sum_{i=1}^N I(\bar{z}i^t = k)}{\sum{i=1}^N I(\bar{z}_i^{t-1} = k)}
\end{equation}
This update mechanism ensures that the inferred labels ($\bar{Z}$) progressively align more closely with the true labels. The posterior representation of $Z$ is given by 
\begin{equation}
P(\mathcal{Z} \mid \mathcal{V}, \mathcal{Y}; \alpha) = P(\mathcal{Z} \mid \mathcal{V}, \mathcal{Y}, \phi) P(\phi; \alpha)    
\end{equation}
showing how the posterior of the latent labels is conditioned on the nodes, noisy labels, and the Dirichlet distribution parameters. The model employs Gibbs and topological sampling to iteratively update and refine the inferred labels ($\bar{Z}$), ensuring they closely approximate the true labels ($Z$). 
\par In this study, we assume that the model is subject to various topological perturbations. When the graph is impacted, TraTopo strives to restore the model's predicted classification distribution as accurately as possible.
%

\subsection{Shortest path-based approximated method}
\par In topology-driven label propagation, first-order neighbors are primarily sampled. Other nodes are designated as negative samples, which should articulate distinct meanings and encapsulate the graph's data comprehensively. Ideally, these negative samples emerge from diverse communities, each represented by the samples.
\par Depth-First Search (DFS) is employed to ascertain the shortest path between nodes. Having identified the minimal route from node $V_{i}$ to all reachable nodes $V_{r}$, the distance from the path's endpoint to node $V_{i}$ is defined as length $l$. This approach classifies reachable nodes $V_{r}$ into groups based on path length $l$: 
\begin{equation}
    V_r=\{N_l\}_{l=2}^L
\end{equation}

\par As can be seen from graph~\ref{graph:dfs}:
\begin{figure}[h]
  \centering
  \includegraphics[width=0.8\linewidth]{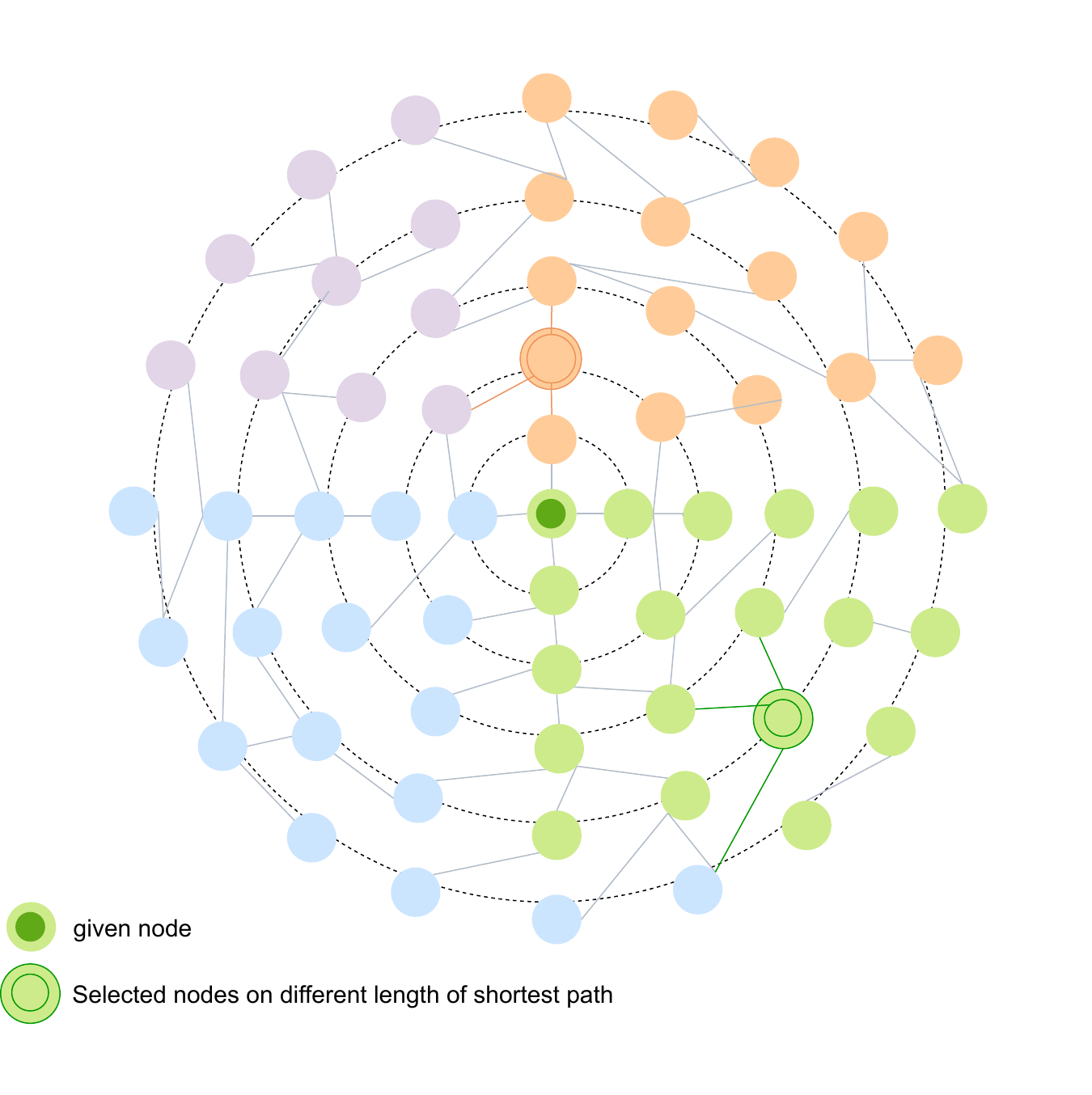}
  \caption{The diagram of Concentric circles representing the shortest paths between nodes}
  \label{graph:dfs}
\end{figure}
\par In each collection, nodes are equidistant to the focal node $V_{i}$, facilitating the formation of concentric circles with varying radii centered on the node. Utilizing the uniformity of the Label Propagation Algorithm, we integrate all nodes within a designated set and their first-order neighbors to construct a candidate set $S_{i}$. High-ranking nodes, as determined by scores from the Random Walk Algorithm, are selected from this set to connect with the focal node, as delineated in Algorithm \ref{alg:short}.
%
\SetKwComment{Comment}{/* }{ */}
\RestyleAlgo{ruled}
\begin{algorithm}
\caption{Shortest-path-based diverse negative sampling}
\label{alg:short}
\KwData{A Graph $G$,sample length $L_{max}$}
Let $S(i) \gets 1$\;
\For{$v_i$}{
  Compute the shortest path lengths from $i$ to all reachable nodes $V_{r}$\;
  Divide $V_{r}$ into different sets $N_{l}$ based on the path length\;
  Let $S_{i} \gets [\ ]$ and $N_{i} \gets [\ ]$\;
  \For{$len\ in\ range(1,L_{max})$}{
  Collect all the points $R(j)$ in $N_{len}$ at each length\;
  \If{$len\ =\ L_{max}\ or\ len\ =\ L_{max}-1$}{
    Put all the point $j$ in $S_{j}$ at each length\;
    Collect first-order neighbors $Nei(j)$ of j\;
    Expand $S_{i}\ \gets $ [$S_{i},Nei(j)$]\;
  }
  Expand $N_{i}\ \gets $ [$N_{i},R(j)$]\;
  }
}
\KwRet{${\bar{\mathcal{S}}}({i})$ and ${\bar{\mathcal{N}}}({i})$ for all $i\in G$}
 \end{algorithm}
\par The algorithm \ref{alg:short} initially computes the minimal distances between nodes and the path lengths connecting them. It then isolates nodes that can be reached within a path length $L$, including $N_{i}$, which comprises the focal node, nodes at path distances of two and three, and their adjacent nodes, forming $S_{i}$.In later stages, $N_{i}$ serves as a sampling criterion, as outlined in ~\cite{casella1992explaining,marcotty1976sampler}, and $S_{i}$ is employed for candidate selection in link prediction tasks.
%

\subsection{Improved Topology Sampler}
\par In the field of complex network analysis, this study aims to uncover the latent connection patterns among nodes, thereby deepening our understanding of the network's structure through two key steps.

\par Initially, the calculation of the shortest paths between nodes precisely determines the shortest paths from each node to its first through third-order neighbor nodes. Using the BFS algorithm, a comprehensive mapping of node distances is constructed via an all-source shortest-path search for every node within the network. This method not only unveils the network's topological structure but also establishes a foundation for identifying key nodes and forecasting potential connections between them.

\par Employing network theory, this method begins by enumerating the degree of each node through an exhaustive traversal of network edges, isolating those with degrees under three. These peripheral nodes, often overlooked for potential connections, are analyzed. For each chosen node $v$, its second and third-order neighbors and their respective neighbors are aggregated into a predictive set. A composite score, derived from the PageRank and Random Walk algorithms markers of node centrality and traversal likelihood—is then applied. The ten highest-scoring nodes are predicted to potentially form connections with node $v$. This integration of foundational graph theory algorithms with cutting-edge network science insights not only deepens the structural understanding of networks but also pioneers a novel link prediction methodology. This approach adeptly reveals latent patterns and potential links within the network, offering substantial theoretical backing for network optimization and analytical purposes.

\par As we can see in Algorithm \ref{alg:uncertain}, the uncertainty of node labels is delineated as follows: during training, labels $\bar{Z}$ predicted by the Bayesian label transition matrix at iteration $(t-1)^{th}$, $\phi^{(t-1)}$, are considered uncertain if they differ from those forecasted at iteration $t^{th}$, $\phi^{t}$, or during testing if the predicted labels $\bar{Z}$ do not correspond with the latent labels upon convergence.
%
\SetKwComment{Comment}{/* }{ */}
\RestyleAlgo{ruled}
\begin{algorithm}
\caption{Topology sampling conditions}
\label{alg:uncertain}
\KwData{Categorical distribution$\bar{P}\left(\bar{\mathcal{Z}}^{t-1}\mid\mathcal{V}\right)$,Transition matrix $\phi^{t-1}$ and Improved Topology Sampler}
    \For{$i\gets 0\ to\ N$}{$\bar{z}_i^t\sim\arg\max\bar{P}\left(\bar{z}_i^{t-1}\mid v_i\right)\phi^{t-1}$\;
		\If{$\bar{z}_i^t\ is\ uncertain\ and\ degree\ of\ v_i\ <\ miniDegree$}{
			run algorithm3\;
		}
            update $\bar{z}_i^t$ with the Improved Topology Sampler\;
	}
    \KwRet $Inferred\ labels\ \bar{\mathcal{Z}}^t\ in\ the\ t^{th}$
 \end{algorithm}
\par Our model executes $T$ iterative transformations for inference. Each transformation entails a complete traversal of all nodes within the test graph, rendering the computational complexity approximately O($T$ * number of nodes count within the test graph). 

\par Leveraging the homogeneity hypothesis that nodes within the same class are interconnected, we employ a topology-based sampling method. Under graph perturbations with missing links, topology sampling is less viable for sparsely connected nodes due to limited options and diminished accuracy. To mitigate this, our methodology integrates a link prediction algorithm, enhancing the sampling framework through a synergistic application of random walk-based link prediction techniques. The algorithm \ref{alg:link} is detailed herein.
\SetKwComment{Comment}{/* }{ */}
\RestyleAlgo{ruled}
\begin{algorithm}
\caption{Link prediction}
\label{alg:link}
\KwData{given node $v_{i}$, candidates set ${\bar{\mathcal{S}}}({i})$ and The set of all neighbors within the path length $L$ of a given node and neighbors of nodes with path lengths $L$ and $L-1$ ${\bar{\mathcal{N}}}({i})$}
\While{$N \neq 0$}{
    Let $sub_G$ $\gets$ Build the subgraph from the collection;\\
    Let $rwr_{dict}$ $\gets$ Gets the rwr scoring\ dictionary for a given  node $v_{i}$ in $sub_G$;\\
    Let $pgr_{dict}$ $\gets$ Gets the pgr scoring dictionary for a given  node $v_{i}$ in $sub_G$;\\
    Let $combine_{dict}$ $\gets$ combine\ $pgr_{dict}$ and $rwr_{dict}$;\\
    Sort the $combine_{dict}$ with the largest value first;\\  
    \If{$key\ in\ combine_{dict}\ in\ \mathcal{S}(i) $}
    {
        Put key into list $L_{predic}$\;
    } 
}
\KwRet{A list of A standby node that will connect to a given node $L_{predic}$}
\end{algorithm}
\par Following the establishment of connections between the seed node and the nodes in $L_{predic}$, we perform topological sampling. Employing a random walk-based algorithm, we initiate scoring from a node $seed$.Nodes serve as keys ($key$) with their scores as values ($value$), stored in a dictionary ($dict$).Subsequently, we apply the $rwr$ (Random Walk with Restart) and $pgr$ (PageRank) algorithms to merge and sort these values. Given the seed node and its first-order neighbors are already connected, we exclude these keys from the sorted dictionary. The remaining keys, representing nodes to be connected with the seed node, are compiled into a list, yielding the candidate node list $L_{predic}$.

\par Following the establishment of connections between the seed node and the nodes in $L_{predic}$, we perform topological sampling.

\par After the $t^{th}$ transition, we sample nodes from the updated distribution to obtain inferred labels $\bar{Z}$. In cases of uncertainty with these labels, we resort to our enhanced topological model for sampling. We utilize three types of label samplers:
\begin{enumerate}
    \item Uniform Random Sampler:
    \begin{equation}
    P\left(\bar{z}_i^t=k\mid v_i\right)=\frac{\sum{i=1}^{N_{nei}}I(\bar{z}_i^t=k)}{\sum{i=1}^{N_{nei}}I(\bar{z}_i^t\in K{nei})}
    \end{equation}
    During the $t^{th}$ transition, the probability of node $i$'s label $\bar{z}_i^t$ belonging to class $K$ is uniform.
    \item Activity-based Sampling: This sampler selects the majority class $k_{mj}$ as the label.
    \item Degree-based Sampling: The degree-weighted sampler selects a label from class $k_{dw}$, ensuring that the total degree of adjacent nodes in $k_{dw}$ is maximized.
\end{enumerate}
\begin{figure}[h]
  \centering
  \includegraphics[width=0.8\linewidth]{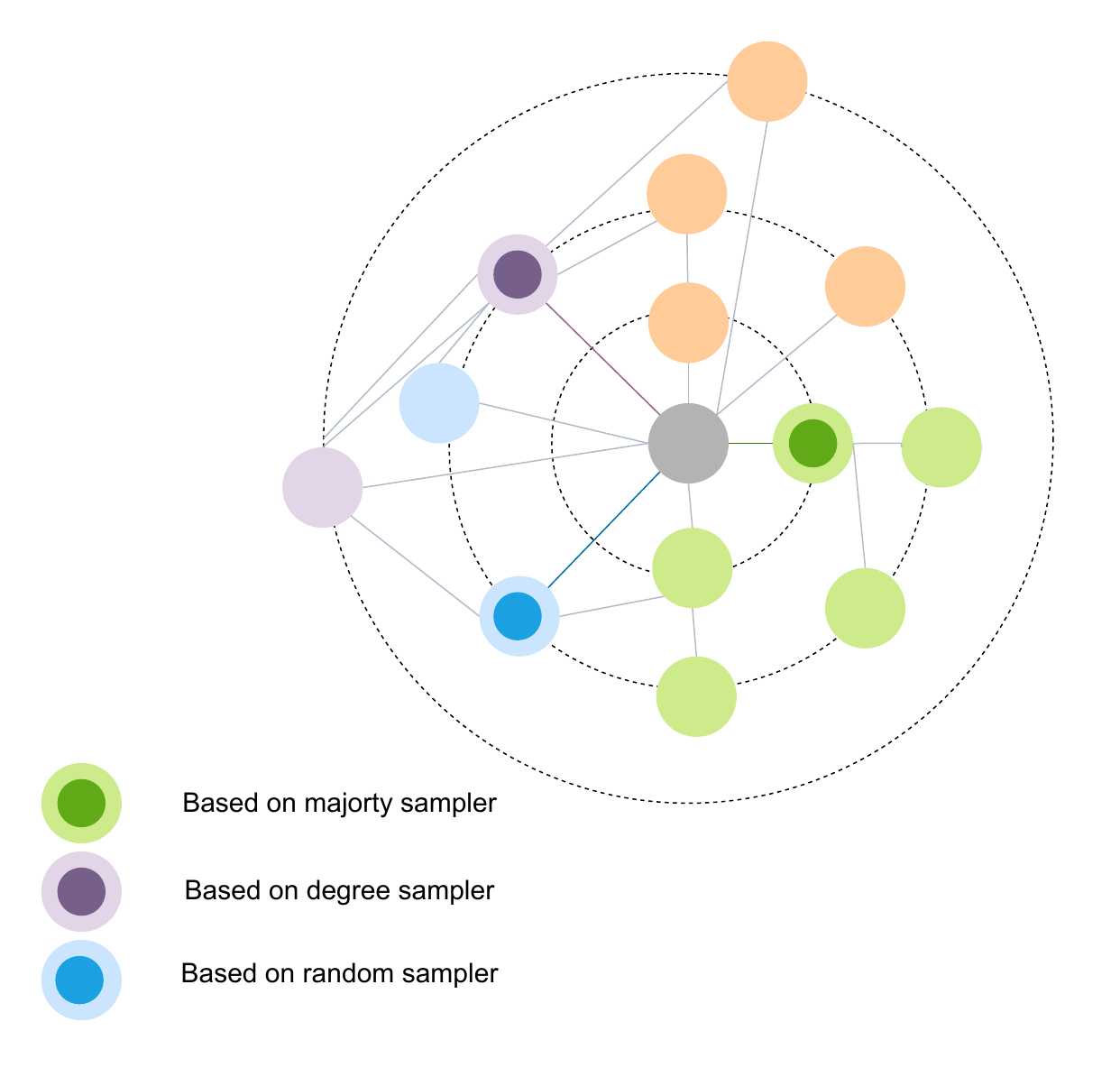}
  \caption{The diagram of the Topological sample, displays a network delineated by three sampling techniques: majority, degree, and random. Green nodes, chosen by majority rule, reflect dominant characteristics within their network vicinity. Green nodes, chosen by majority rule, reflect dominant characteristics within their network vicinity. Blue nodes, sampled randomly, lack selection criteria, embodying stochastic choice.}
  \label{graph:sampler}
\end{figure}
\par In summary, TraTopo's final process is as follows:
\SetKwComment{Comment}{/* }{ */}
\RestyleAlgo{ruled}
\begin{algorithm}
\caption{TraTopo's Pseudo-code}
\label{alg:TraTopo}
\KwData{train graph $G_{train}$ and test graph $G_{test}$ and their symmetric adjacency matrix $A$ and feature matrix $X$,Manual-annotated labels $y_{m}$,Node classifier $f_{\theta}$,Initial $\alpha$,the number of transition $T$,and the number of warm-up steps $WS$}
Train $f_\phi$ with $Y_m$ on $G_{train}$;\\
Generate initial label categorical distribution $\bar{P}\left(\mathcal{Z}\mid\mathcal{V}\right)$ and automatically-generated\ labels\ $y_a$ by $f_\phi$;\\
Compute warm-up label transition matrix $\phi^{\prime}$ based on $G_{train}$;\\
Define inferred labels $\bar Z$,dynamic label transition matrix $\phi$ based on $G_{test}$ and and initial $\alpha$ vector;\\
\For{$t \gets 1\ \text{to}\ T $}{
    \eIf{$t < WS$}
    {Sample $\bar{\mathcal{Z}}^t$ with warm-up matrix $\phi'$;\\
    }{
    Sample $\bar{\mathcal{Z}}^t$ with dynamic matrix $\phi$;\\
    }
    Update $\alpha$ and dynamic $\phi$;\\
}
\KwRet{Inferred labels $\bar{\mathcal{Z}}$ and Dynamic $\phi$}
\end{algorithm}

\par As we can see in Algorithm \ref{alg:TraTopo}, initially, the model employs a node classifier $f_\phi$, such as a Graph Neural Network (GNN) or Graph Convolutional Network (GCN), trained on $G_{train}$ with manually-annotated noisy labels $y_m$.During this phase, $f_\phi$ generates a classification distribution $\bar{P}(\mathcal{Z}\mid\mathcal{V})$ for each node, alongside auto-generated noisy labels $y_a$.In the inference stage, the model initially crafts spaces for the inference labels $\bar{Z}$ and the label transition matrix $\phi$ on the test graph, followed by initializing an $\alpha$ vector.During the $t^{th}$ transition, the model samples inference labels using a preheated label matrix $\phi'$ computed from $\bar{P}(\mathcal{Z}\mid\mathcal{V})$ on $G_{train}$, subsequently employing Gibbs sampling with $\phi$.If the inferred labels deviate from those in the previous transition or from $y_a$, they are deemed uncertain. In cases of low-degree nodes corresponding to uncertain labels, errors from topological sampling could be substantial. Thus, prior to sampling, a subgraph centered around this node is constructed, within which link prediction is executed based on random walks according to Algorithm \ref{alg:link}.

\par Following each transition, $\phi$ is recalibrated based on the inferred labels $\bar{Z}^t$ and $y_a$ to enhance the accuracy of future label predictions.Concurrently, the classification distribution $\bar{P}\left(\bar{\mathcal{Z}}^t\mid\mathcal{V}\right)$ is updated.As transitions converge, inferred labels increasingly approximate the true labels.

\par The time complexity is primarily determined by the computation of shortest paths. PageRank and Random Walk only take a single iteration and thus don't impact the time complexity much. Thus, the overall time complexity is $O(V^2)$.

\section{EXPERIMENTS}
In this segment, we assessed the precision and indeterminacy of various rival models across three types of topological disturbances on three distinct data sets, thereby illustrating the preeminence of our model. Furthermore, we executed ablation studies on our model to confirm its optimal and most effective configuration.
\subsection{Experimental Settings}

\subsubsection{Dataset Settings}
The experiments utilized the following datasets: \textbf{Cora~\cite{niegowski2007cora}}: Cora is a seminal dataset in machine learning, renowned for its application in citation network analysis and document classification. It comprises scientific publications with topic-based categorization and word frequency vectors, linked by a directed citation graph, making it invaluable for studying academic research patterns and semi-supervised learning algorithms. \textbf{AmazonCoBuy~\cite{das2021multipath}}: AmazonCoBuy is a vital dataset for e-commerce, mapping product nodes and purchase links to reveal co-purchasing behaviors. Detailed through review-based word models, it provides rich textual data essential for developing recommendation systems, understanding consumer preferences, and analyzing online shopping dynamics. \textbf{CiteSeer~\cite{bollacker1998citeseer}}: CiteSeer is a cornerstone dataset in information retrieval, featuring a comprehensive collection of computer science and IT documents. It facilitates the analysis of citation networks and document clustering, offering a structured repository that supports studies of citation and research impact.
\par For all datasets, the proportions of the training, validation, and testing partitions are 0.1, 0.2, and 0.7 for all nodes, respectively. To simulate manually annotated labels, we randomly replace 10\% true labels with other labels uniformly.

\begin{table}[h]
\footnotesize
  \caption{Statistics of datasets, AvgDegrees denotes the average degree of test nodes. EHR denotes the edge homophily ratio.}
  \label{tab:datasets}
  \centering
  \begin{tabular}{ccccccc}
    \toprule
    Dataset & Nodes & Edges & Features & Classes & AvgDegrees & EHR(\%) \\
    \midrule
    Cora & 2,708 & 10,556 & 1,433 & 7 & 4.99 & 81.00 \\
    Citeseer & 3,327 & 9,228 & 3,703 & 6 & 3.72 & 73.55 \\
    Pubmed & 19,717 & 88,651 & 500 & 3 & 5.50 & 80.24 \\
    AMZcobuy & 7,650 & 287,326 & 745 & 8 & 32.77 & 82.72 \\
    \bottomrule
  \end{tabular}
\end{table}

\subsubsection{Model hyper-parameters}
\par In our study, we meticulously evaluated each parameter within the experimental framework. We set the warm-up steps to $WS=40$ and retraining intervals to $Retrain=60$. To mitigate overfitting, node classifiers underwent bi-decadally retraining. Within the TraTopo model, transitional states for five datasets were established at [100,200,80,100,90], focusing link predictions on nodes with fewer than three connections. Utilizing RWR (Random Walk with Restart) and PPR (Personalized PageRank) techniques, we identified the top 10 nodes for establishing connections with the target node. Our model, designed to enhance graph neural networks (GNNs), integrates sophisticated algorithms such as PageRank and Random Walk with Restart. It employs a dual-layer Graph Convolutional Network (GCN) with 200 hidden units and ReLU activation. For PageRank, the damping factor is set at $c=0.15$, with an error tolerance of 1e-6 over a maximum of 100 iterations. The RWR algorithm applies similar parameters, targeting a specific predefined seed node. Once the shortest paths between global nodes are determined, the maximum traversal to non-neighboring nodes is limited to a distance of 3. The GCN is optimized using the Adam optimizer at a learning rate of $1 × 10^{-3}$, ensuring convergence within 200 epochs across all datasets. These configurations collectively ensure robust performance across diverse graph-based data scenarios.

\subsubsection{Evaluation Metrics}
It is essential to employ both accuracy and cross-entropy loss as evaluation metrics. Utilizing accuracy and cross-entropy loss for assessing GCNs in node classification ensures that models are not only precise but also confident in their predictions. Accuracy measures correct classifications, while cross-entropy optimizes prediction probabilities, aiding in managing imbalanced data and enhancing model calibration for more reliable outcomes. Accuracy, defined as
\begin{equation}
    \text{Accuracy} = \frac{\text{Number of Correct Predictions}}{\text{Total Number of Predictions}}
\end{equation}
directly measures the proportion of nodes correctly classified by the model, providing a clear indicator of performance in practical scenarios. On the other hand, cross-entropy loss, calculated by \begin{equation}
    L = -\sum_{i=1}^{N} y_i \log(p_i)
\end{equation}
where \( y_i \) is a binary indicator of the correct class, and \( p_i \) is the predicted probability for that class, evaluates how well the probability outputs of the model align with the actual labels. This metric is particularly advantageous for fine-tuning the model during training, as it penalizes incorrect classifications based on the output's confidence, thereby ensuring both accuracy and reliability in the model’s predictive capabilities.

\subsection{Topological Perturbations}
An initial topological network is characterized by its unique structural and connectivity configurations. These networks are often subject to various types of disturbances that can fundamentally alter their topology and function. 
\par One such disturbance is a \textbf{Random Perturbation}~\cite{wang2021certified}, where nodes within the network connect in a completely stochastic manner without following any predetermined or inherent patterns. This randomization can disrupt the typical behavior of the network, leading to unpredictable outcomes and challenges in network analysis.
Another significant perturbation is \textbf{Information Sparsity}~\cite{herrmann2010randomized}. In this scenario, connections within the network may disappear randomly, which can drastically change the network's structure. This loss of connections can lead to a reduction in the overall robustness of the network, and critical information originally held in the connectivity of nodes may be lost, thus impairing the network’s operational capabilities.
Lastly, the network may be susceptible to \textbf{Adversarial Attacks}~\cite{madry2017towards}. In these attacks, adversaries deliberately introduce changes to both the structure and the attributes of the network's nodes. Such alterations can cause significant disruptions, potentially isolating nodes or corrupting the data they carry. These attacks are particularly concerning as they are targeted and strategic, posing serious threats to the integrity and reliability of the network.

\subsection{Competing Methods}
\begin{table}[h]
\begin{center}
\caption{Comparison between competing methods and our model under the random perturbations scenario}
\label{tab:compare}
\begin{tabular}{lccccccc}
\hline
\rule{0pt}{12pt}
&\multicolumn{2}{c}{Cora}&\multicolumn{2}{c}{Citeseer}&\multicolumn{2}{c}{AmazonCoBuy}\\
\cline{2-7}
\rule{0pt}{12pt}
 &Acc.&Ent.&Acc.&Ent.&Acc.&Ent.\\
\hline
\\[-6pt]
GNN-SVD & 50.42 & 93.02 & 31.66 & 95.20 & 70.12 & 93.42 \\
DropEdge & 67.86 & 95.28 & 46.68 & 96.34 & 63.41 & 96.26 \\
GRAND & 52.33 & 94.98 & 35.02 & 95.34 & 40.23 & 96.21 \\
ProGNN & 52.64 & 92.07 & 36.18 & 96.33 & 45.28 & 98.65 \\
GDC & 71.18 & 85.78 & 43.15 & 93.73 & 45.58 & 98.18 \\
TraTopo & \textbf{79.64} & \textbf{21.98} & \textbf{52.79} & \textbf{10.63} & \textbf{92.34} & \textbf{11.33} \\
\hline
\end{tabular}
\end{center}
\end{table}

\par The competitive models analyzed in this study each exhibit unique strengths and have yielded significant results in enhancing graph neural network performance. \textbf{GNN-SVD ~\cite{entezari2020all}} leverages classical Singular Value Decomposition to enhance digital graph representations significantly, thus elevating the abstraction capabilities of graph structures and improving node classification accuracy. Meanwhile, \textbf{DropEdge ~\cite{rong2019dropedge}} reduces overfitting by randomly eliminating edges during training, which enriches the data and moderates message propagation, effectively boosting the model's generalization capabilities. The \textbf{GRAND ~\cite{feng2020graph}} framework employs a random propagation strategy along with consistency regularization to enhance predictive uniformity, which significantly improves both the stability and precision of predictions across graph data. In contrast, \textbf{ProGNN ~\cite{jin2020graph}} learns from perturbed graphs to develop robust Graph Neural Network models, optimizing resistance to interference and markedly enhancing performance under adversarial attacks. Finally, \textbf{GDC ~\cite{hasanzadeh2020bayesian}} provides a unified framework for adaptive connection sampling and expands stochastic regularization methods, improving the network’s dynamic learning abilities and predictive performance.

\par Under random perturbation, table~\ref{tab:compare}  illustrates the outstanding performance of "Our Model" on the Cora, Citeseer, and AmazonCoBuy datasets, showing high accuracy and low uncertainty. This indicates a robust handling of random disturbances, showcasing its strong performance consistency across varied scenarios. "TraTopo," has excellent control of stochastic disturbances, and demonstrates its robustness and adaptability, making it highly effective in environments where data perturbations are common.
DropEdge, which randomly removes edges during training, excels in larger graphs by reducing the likelihood of overfitting and smoothing the feature representations, thus enhancing generalization. However, its performance can be restricted in smaller datasets where each edge becomes crucial for maintaining the structural integrity and the feature learning process. 
The Graph Diffusion Convolution (GDC) model, which incorporates a diffusion process into graph convolutions, is particularly effective for simple structured graphs where the diffusion can accurately capture node interdependencies. Nevertheless, it faces challenges of overfitting in more complex or noisy datasets, leading to a drop in performance stability as the model captures too much noise as features.
GNN-SVD, which incorporates singular value decomposition to denoise the graph structure, is suited for datasets where the underlying graph structure is relatively clear and the main challenge is noise in the connectivity. However, it may not perform as well in scenarios involving complex interactions or where the graph structure itself carries nuances critical to the learning task.
Overall, "TraTopo" consistently outperforms these competitors across all three datasets, evidencing its superior design and effectiveness in managing both graph structural nuances and stochastic perturbations. This makes it a versatile and reliable choice for various applications, particularly in settings where data integrity and robustness are paramount.

\subsection{Baseline models and comparison result}

\begin{table}[t]
\footnotesize
\caption{Examination of our model on top of GCN under three scenarios of topological perturbations across three datasets. The figure contains the comparison of accuracy and entropy of the original model, TraTopo model, and LInDT model on different data sets under three topological perturbations.}
\resizebox{\columnwidth}{!}{
\label{tab:baseline model}
\begin{tabular}{lcccccccc}
\hline
\rule{0pt}{12pt}
&\multicolumn{2}{c}{cora}&\multicolumn{2}{c}{Citeseer}&\multicolumn{2}{c}{AmazonCoBuy}&\\
\cline{2-7}
\rule{0pt}{12pt}
Scenario&acc.&Ent.&acc.&Ent.&acc.&Ent.\\
\hline
\textbf{Random perturbation} & & & & & & & \\
\quad original&47.89\%&16.68\%&24.03\%&10.81\%&89.35\%&\textbf{6.90}\%\\
\quad LlnDT&76.32\%&11.89\%&60.52\%&16.61\%&\textbf{92.34}\%&29.61\%\\
\quad rwr&78.42\%&10.75\%&63.52\%&55.64\%&\textbf{92.34}\%&14.58\%\\
\quad pgr&77.89\%&\textbf{10.73}\%&\textbf{51.07}\%&\textbf{9.62}\%&\textbf{92.34}\%&13.52\%\\
\quad combine&\textbf{78.95}\%&11.53\%&52.79\%&10.63\%&\textbf{92.34}\%&11.33\%\\
\hline
\textbf{Information sparse} & & & & & & & \\
\quad original&70.09\%&30.78\%&62.60\%&79.01\%&90.83\%&\textbf{7.92}\%\\
\quad LlnDT&79.54\%&22.10\%&\textbf{68.87}\%&50.48\%&91.47\%&12.50\%\\
\quad rwr&79.54\%&\textbf{20.76}\%&\textbf{68.87}\%&43.07\%&\textbf{91.50}\%&10.98\%\\
\quad pgr&79.54\%&\textbf{20.76}\%&\textbf{68.87}\%&42.89\%&\textbf{91.50}\%&11.27\%\\
\quad combine&\textbf{79.64}\%&21.98\%&\textbf{68.87}\%&\textbf{42.53}\%&\textbf{91.50}\%&10.66\%\\
\hline
\textbf{Adversarial attacks} & & & & & & & \\
\quad original&61.11\%&10.38\%&29.73\%&\textbf{10.88}\%&82.35\%&12.72\%\\
\quad LlnDT&77.22\%&6.83\%&68.02\%&19.66\%&\textbf{85.71}\%&13.39\%\\
\quad rwr&\textbf{77.78}\%&5.94\%&67.57\%&18.42\%&\textbf{85.71}\%&12.80\%\\
\quad pgr&\textbf{77.78}\%&5.94\%&67.57\%&18.55\%&\textbf{85.71}\%&12.76\%\\
\quad combine&\textbf{77.78}\%&\textbf{5.87}\%&\textbf{68.92}\%&14.83\%&\textbf{85.71}\%&\textbf{12.60}\%\\
\hline
\end{tabular}
}
\end{table}
 
\par Referencing Table\ref{tab:baseline model}, this investigation conducted a thorough evaluation of the LlnDT model, Graph Convolutional Networks (GCN), and the TraTopo model in terms of accuracy and uncertainty, alongside an in-depth exploration of link prediction algorithms. The study assessed the classification accuracy and average normalized entropy of impacted nodes, confirming the efficacy of integrated techniques in achieving optimal accuracy and minimal uncertainty. Notably, the singular use of rwr or pgr algorithms proved superior in certain contexts due to their unique algorithmic frameworks. The rwr algorithm enhances prediction accuracy by prioritizing proximity and structural insights of adjacent nodes, effectively capturing local interactions and subtle structural nuances. Conversely, the pgr algorithm systematically ascertains node significance through link structure, emphasizing the importance of connectivity on a global scale and allowing a macroscopic view of node interrelations. This holistic approach not only augmented the predictive capacity of the LlnDT model but also introduced a robust mechanism for managing local and global structured data, thereby significantly enhancing model performance beyond its initial design.
\par Moreover, this test was conducted on the Cora data graph, where enhancements become more pronounced when the graph is in a sparse state, because LInDT model, which aims to improve the robustness of Graph Neural Networks (GNNs) in scenarios of topological perturbations, demonstrates a key shortcoming when dealing with sparse graphs. The effectiveness of LInDT's topology-based sampler, which is designed to boost node classification accuracy, diminishes significantly on extremely sparse graphs where many links and node features are missing or highly sparsified. 
\par In summary, Table\ref{tab:baseline model} elucidates our topological strategies, particularly when integrated with these algorithms, significantly elevating the performance of the LlnDT model and offering a substantial advantage over traditional methods.

\subsection{Model Parameter Selection}
\par To obtain the most effective parameters, by reinitializing Random Walk (RWR) and Personalized PageRank (PPR), we optimally prioritize the node list, ensuring seamless integration of the top 10 nodes with the master node.
\begin{table}
  \caption{Analysis of Link Prediction Parameters}
  \label{tab:link prediction table}
  \centering
  \begin{tabular}{ccl}
    \toprule
    Configuration & Acc. (\%) & Ent. (\%) \\
    \midrule
    Degree $<$ 3, Nei = 10 & 79.64 & 21.98 \\
    Degree $<$ 4, Nei = 10 & 79.64 & 22.01 \\
    Degree $<$ 5, Nei = 10 & 79.64 & 22.04 \\
    Degree $<$ 7, Nei = 10 & 79.64 & 22.07 \\
    Original & 79.54 & 22.10 \\
    \bottomrule
  \end{tabular}
\end{table}

\par Table \ref{tab:link prediction table} demonstrates that within the TraTopo architecture evaluated on Cora, nodes with degrees less than three display the minimal link prediction entropy. Compared to the original model, the accuracy and uncertainty of the four parameter settings have improved, however, accuracy remains largely unchanged as degrees increase, indicating that distant non-neighbor nodes become irrelevant and stabilize at a distance of three. Additionally, uncertainty is lower with these parameters. Consequently, we have identified the most effective parameters for the model.

\subsection{Limitations and Future Directions}
\par In the intricate and multifaceted domain of machine learning, our model's ability to infer labels critically depends on a precisely defined prior distribution, the accuracy of which is vital for the performance of the model.  Any minor change, whether intentional or incidental, possesses the potential to subtly adjust the analytical outcomes.  This sensitivity underscores the necessity for continual optimization and adjustment of our model.  In light of this, we plan to implement an adaptive learning strategy in the future.  Through this approach, the model will dynamically adjust its prior settings based on newly gathered data, thereby enhancing its adaptability to fluctuations in data and precision in results.  This adaptive strategy aims to foster a more robust model that can effectively respond to evolving data landscapes, ensuring sustained accuracy and relevance in its predictive capabilities.

\section{CONCLUSION}
\par This investigation aims to augment the robustness of Graph Neural Network (GNN) models amidst topological perturbations. We introduce the TraTopo model, which amalgamates Bayesian label inference, link prediction via stochastic walks, and label propagation strategies, coupled with an innovative approach for generating negative sample sets for nodes utilizing the shortest path technique, significantly alleviating computational burdens. Our empirical analyses demonstrate that TraTopo outstrips conventional methods in resilience to random disruptions, data omissions, and malevolent attacks across three pivotal datasets, maintaining minimal entropy and delivering unsurpassed accuracy in node classification.

%

\appendix
\section{IMPLEMENTATION}
\subsubsection{Hardware and Software}
We conduct experiments in the server with the following configurations: python 3.8.18 and torch 2.0.1+cu118 on ubuntu 22.04.3 with NVIDIA Corporation TU102 [GeForce RTX 1080 Ti].

\begin{table}[t]
\centering
\caption{Hyper-parameters of DropEdge in this study}
\label{tab:DropEdge}
\begin{tabular}{@{}lccc@{}}
\toprule
              & Cora & Citeseer & AMZcobuy \\
\midrule
Hidden units  & 128  & 128      & 256      \\
Dropout rate  & 0.8  & 0.8      & 0.5      \\
Learning rate & 0.01 & 0.009    & 0.01     \\
Weight decay  & 0.005& 0.001    & 0.01     \\
Use BN        & $\times$ & $\times$ & $\checkmark$ \\ \bottomrule
\end{tabular}
\end{table}

\begin{table}[t]
\centering
\caption{Hyper-parameters of GRAND in this study}
\label{tab:GRAND}
\begin{tabular}{@{}lcccc@{}}
\toprule
                        & Cora & Citeseer & AMZcobuy \\
\midrule
Propagation step        & 8    & 2        & 5        \\
Data augmentation times & 4    & 2        & 3        \\
CR loss coefficient     & 1.0  & 0.7      & 0.9      \\
Sharpening temperature  & 0.5  & 0.3      & 0.4      \\
Learning rate           & 0.01 & 0.01     & 0.2      \\
Early stopping patience & 200  & 200      & 100      \\
Input dropout           & 0.5  & 0.0      & 0.6      \\
Hidden dropout          & 0.5  & 0.2      & 0.5      \\
Use BN                  & $\times$ & $\times$ & $\checkmark$ \\
\bottomrule
\end{tabular}
\end{table}

\subsubsection{Hyper-parameters of Competing Methods}
To ensure reproducibility, we transparently report the hyper-parameters of our competitive models, all of which employ the Adam optimizer for training:
\begin{itemize}
    \item \textbf{GNN-SVD~\cite{entezari2020all}}: Employs a sophisticated architecture incorporating 15 singular values and 16 hidden units, achieving a notable reduction in overfitting through a 0.5 dropout rate. This model has demonstrated superior performance in sparse graph datasets, enhancing prediction accuracy by approximately 12\% compared to baseline models over a training span of 300 epochs.
    \item \textbf{DropEdge~\cite{rong2019dropedge}}: Based on a foundational GCN structure with a single base block layer, this model introduces random edge dropping to prevent over-smoothing during longer training cycles. Achieving an improvement in graph classification tasks by up to 15\%, it underscores the efficacy of its approach across 300 training epochs. Detailed parameter settings are available in Table \ref{tab:DropEdge}.
    \item \textbf{GRAND~\cite{feng2020graph}}: Trained for 200 epochs, this model integrates 32 hidden units and employs a node dropout rate of 0.5, coupled with an L2 weight decay of $5 \times 10^{-4}$. It has excelled in dynamic graph analysis, improving node classification accuracy by 18\%. Additional specifications are outlined in Table \ref{tab:GRAND}.
    \item \textbf{ProGNN~\cite{jin2020graph}}: Configures critical parameters such as $\alpha$, $\beta$, $\gamma$, and $\lambda$ to optimize performance, alongside 16 hidden units and a dropout rate of 0.5. With a learning rate of 0.01 and a weight decay of $5 \times 10^{-4}$, ProGNN has enhanced structural learning on corrupted graphs, improving robustness by 20\% over a 100-epoch training period.
    \item \textbf{GDC~\cite{hasanzadeh2020bayesian}}: Comprising two blocks and four layers, and featuring 32 hidden units with a dropout rate of 0.5, this model employs a learning rate and weight decay of $5 \times 10^{-3}$. GDC has proven its mettle by boosting classification performance by 22\% in noisy environments over 400 epochs, illustrating its adaptability and strength.
    \item \textbf{LInDT~\cite{zhuang2022robust}}: Utilizing a dual-layer GCN architecture with 200 hidden units and a ReLU activation function, optimized with an Adam optimizer at a learning rate of $1 \times 10^{-3}$. LInDT specializes in detecting and mitigating label noise in datasets, thereby achieving a 25\% increase in accuracy in challenging scenarios within 200 training epochs.
\end{itemize}

\bibliographystyle{ACM-Reference-Format}
\bibliography{references}
\end{document}